\documentclass[runningheads]{llncs}
\usepackage{graphicx}
\usepackage{gensymb}
\usepackage{amsmath}
\usepackage{siunitx}
\usepackage{ragged2e}
\usepackage{cite}

\usepackage{tikz}
\usepackage{ctable}
\usepackage{pgfplots}
\pgfplotsset{width=7cm,compat=1.3}
\begin{document}
\title{LYTNet: A Convolutional Neural Network for Real-Time Pedestrian Traffic Lights and Zebra Crossing Recognition for the Visually Impaired}

\titlerunning{LYTNet: Deep Learning Network for the Visually Impaired}
%
\author{Samuel Yu\inst{1} \and
Heon Lee\inst{1} \and
John Kim\inst{1}}

\authorrunning{Yu et al.}
%
\institute{Shanghai American School Puxi West, Shanghai 201103, China
\email{samuel01px2020@saschina.org}\\
\email{heon01px2020@saschina.org}\\
\email{john01px2020@saschina.org}}

\maketitle              
\begin{abstract}
Currently, the visually impaired rely on either a sighted human, guide dog, or white cane to safely navigate. However, the training of guide dogs is extremely expensive, and canes cannot provide essential information regarding the color of traffic lights and direction of crosswalks. In this paper, we propose a deep learning based solution that provides information regarding the traffic light mode and the position of the  zebra crossing. Previous solutions that utilize machine learning only provide one piece of information and are mostly binary: only detecting red or green lights. The proposed convolutional neural network, LYTNet, is designed for comprehensiveness, accuracy, and computational efficiency. LYTNet delivers both of the two most important pieces of information for the visually impaired to cross the road. We provide five classes of pedestrian traffic lights rather than the commonly seen three or four, and a direction vector representing the midline of the zebra crossing that is converted from the 2D image plane to real-world positions. We created our own dataset of pedestrian traffic lights containing over 5000 photos taken at hundreds of intersections in Shanghai. The experiments carried out achieve a classification accuracy of 94\%, average angle error of 6.35\degree, with a frame rate of 20 frames per second when testing the network on an iPhone 7 with additional post-processing steps.

\end{abstract}
\section{Introduction}
The primary issue that the visually impaired face is not with obstacles, which can be detected by their cane, but with information that requires the ability to see. When we interviewed numerous visually impaired people, there was a shared concern regarding safely crossing the road when traveling alone. The reason for this concern is that the visually impaired cannot be informed of the color of pedestrian traffic lights and the direction in which they should cross the road to stay on the pedestrian zebra crossing. When interviewed, they reached a consensus that the information stated above is the most essential for crossing roads.

To solve this problem, some hardware products have been developed \cite{1}. However, they are too financially burdening due to both the cost of the product itself and possible reliance on external servers to run the algorithm. The financial concern is especially important for the visually impaired community in developing countries, such as the people we interviewed who live in China. Accordingly, our paper addresses this issue by discussing LYTNet that can later be deployed on a mobile phone, both ios and android, and run locally. This method would be a cheap, comprehensive, and easily accessible alternative that supplements white-canes for the visually impaired community. 

We propose LYTNet, an image classifier, to classify whether or not there is a traffic light in the image, and if so, what color/mode it is in. We also implement a zebra crossing detector in LYTNet that outputs coordinates for the midline of the zebra crossing.

The main contributions of our work are as follows:
\begin{itemize}
    \item To the best of our knowledge, we are the first to create a convolutional neural network (LYTNet) that outputs both the mode of the pedestrian traffic light and midline of the zebra crossing 
    \item We create and publish the largest pedestrian traffic light dataset, consisting of 5059 photos with labels of both the mode of traffic lights and the direction vector of the zebra crossing \cite{2}
    \item We design a lightweight deep learning model (LYTNet) that can be deployed efficiently on a mobile phone application and is able to run at 20 frames per second (FPS)
    \item We train a unique deep learning model (LYTNet) that uses one-step image classification instead of multiple steps, and matches previous attempts that only focus on traffic light detection

\end{itemize}
The rest of the paper is organized in the following manner: Section II discusses previous work and contributions made to the development and advancements in the detection of pedestrian traffic light detectors and zebra crossings; Section III describes the proposed method of pedestrian traffic light and zebra crossing classifier; Section IV provides experiment results and comparisons against a published method; Section V concludes the paper and explores possible future work.

\section{Related Works}
Some industrialized countries have developed acoustic pedestrian traffic lights that produce sound when the light is green, and is used as a signal for the visually impaired to know when to cross the street \cite{3,4,5}. However, for less economically developed countries, crossing streets is still a problem for the blind, and acoustic pedestrian traffic lights are not ubiquitous even in developed nations \cite{3}. 

The task of detecting traffic light for autonomous driving has been explored by many and has developed over the years \cite{6,7,8,9}. Behrendt et al. \cite{10} created a model that is able to detect traffic lights as small as $3 \times 10$ pixels and with relatively high accuracy. Though most models for traffic lights have a high precision and recall rate of nearly 100\% and show practical usage, the same cannot be said for pedestrian traffic lights. Pedestrian traffic lights differ because they are complex shaped and usually differ based on the region in which the pedestrian traffic light is placed. Traffic lights, on the other hand, are simple circles in nearly all countries.

Shioyama et al. \cite{11} were one of the first to develop an algorithm to detect pedestrian traffic lights and the length of the zebra-crossing. Others such as Mascetti et al. and Charette et al. \cite{3,15} both developed an analytic image processing algorithm, which undergoes candidate extraction, candidate recognition, and candidate classification. Cheng et al. \cite{5} proposed a more robust real-time pedestrian traffic lights detection algorithm, which gets rid of the analytic image processing method and uses candidate extraction and a concise machine learning scheme. 

A limitation that many attempts faced was the speed of hardware. Thus, Ivanchenko et al. \cite{12} created an algorithm specifically for mobile devices with an accelerator to detect pedestrian traffic lights in real time. Angin et al. \cite{13}  incorporated external servers to remove the limitation of hardware and provide more accurate information. Though the external servers are able to run deeper models than phones, it requires fast and stable internet connection at all times. Moreover, the advancement of efficient neural networks such as MobileNet v2 enable a deep-learning approach to be implemented on a mobile device \cite{14}.  

Direction is another factor to consider when helping the visually impaired cross the street. Though the visually impaired can have a good sense of the general direction to cross the road in familiar environments, relying on one's memory has its limitations \cite{16}. Therefore, solutions to provide specific direction have also been devised. Other than detecting the color of pedestrian traffic lights, Ivanchenko et al. \cite{16} also created an algorithm for detecting zebra crossings. The system obtains information of how much of the zebra-crossing is visible to help the visually impaired know whether or not they are generally facing in the correct direction, but it does not provide the specific location of the zebra crossing. Poggi et al., Lausser et al., and Banich \cite{17,18,19} also use deep learning neural network within computer vision to detect zebra crossings to help the visually impaired cross streets. However, no deep learning method is able to output both traffic light and zebra crossing information simultaneously. 

\section{Proposed Method}
Our method is performed on our labeled test-set. The training, test, and validation sets do not overlap.
\begin{figure}
\includegraphics[width=\textwidth]{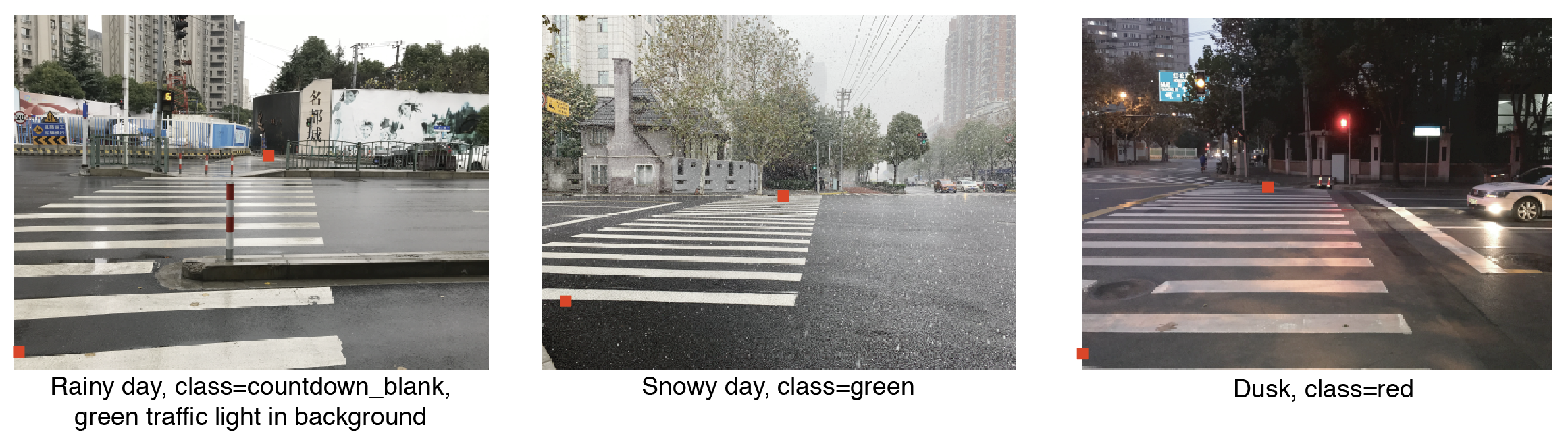}
\caption{Sample images taken in different weather and lighting conditions. Other pedestrian traffic lights or vehicle/bicycle traffic lights can be seen in the images. The two endpoints of the zebra crossing are labelled as seen on the images.} \label{figure1}
\end{figure}
\subsection{Dataset Collection and Pre-Processing}
Our data consists of images of street intersection scenes in Shanghai, China in varying weather and lighting conditions. Images were captured with two different cameras, an iPhone 7 and iPhone 6s at a resolution of $4032 \times 3024$ \cite{2}. The camera was positioned at varying heights and angles around the vertical and transverse axes, but the angle around the longitudinal axis was kept relatively constant under the assumption that the visually impaired are able to keep the phone in a horizontal orientation. At an intersection, images were captured at varying positions relative to the center of the crosswalk, and at different positions on the crosswalk. Images may contain multiple pedestrian traffic lights, or other traffic lights such as vehicle and bicycle traffic lights. 

The final dataset consists of 5059 images \cite{2}. Each image was labelled with a ground truth class for traffic lights: red, green, countdown\_green, countdown\_blank, and none. Sample images are shown in Figure 1. Images were also labelled with 2 image coordinates $(x,y)$ representing the endpoints of the zebra crossing as pictured on the image. The image coordinates define the midline of the zebra crossing. In a significant number of the images, the mid-line of the zebra crossing was obstructed by pedestrians, cars, bicycles, or motorcycles. Statistics regarding the labelled images are shown in Table 1. 

Prior to training, each image was re-sized to a resolution of $876 \times 657$. During each epoch, a random crop of size $768 \times 576$ and a random horizontal flip was applied to each image to prevent over-fitting. The training dataset was partitioned into 5 equal groups and 5-fold cross validation was performed. Images used in the validation dataset were directly re-sized from $4032 \times 3024$ to $768 \times 576$ without any transformations applied. 
\begin{table}
\caption{Composition of Dataset}\label{tab1} 
\begin{tabular*}{\textwidth}{c @{\extracolsep{\fill}} cccccc}
\specialrule{.1em}{.05em}{.05em}
 &  Red & Green & CD Green & CD Blank & None & Total\\
\specialrule{.1em}{.05em}{.05em} 
Number of Images & 1477 & 1303 & 963 & 904 & 412 & 5059\\
\newline
Percentage of Dataset & 29.2\% & 25.8\% & 19.0\% & 17.9\% & 8.1\% & 100.0\%\\
\specialrule{.1em}{.05em}{.05em} 
\end{tabular*}
\end{table}

\subsection{Classification and Regression Algorithm}
Our neural network, LYTNet, follows the framework of MobileNet v2, a lightweight neural network designed to operate on mobile phones. MobileNet v2 primarily uses depthwise separable convolutions. In a depthwise separable convolution, a "depthwise" convolution is first performed: the channels of the input image are separated and different filters are used for every convolution over each channel. Then, a pointwise convolution (regular convolution of kernel size $1 \times 1$) is used to collapse the channels to a depth of 1. For an input of dimensions $h_i \cdot w_i \cdot d_i$ convolved with stride 1 with a kernel of size $k \cdot k$ and $d_j$ output channels, the cost of a standard convolution is $h_i \cdot w_i \cdot k^2 \cdot d_i \cdot d_j$ while the cost of a depthwise separable convolution is $h_i \cdot w_i \cdot d_i \cdot (k^2 + d_j)$ \cite{14}. Thus, the total cost of a depthwise separable convolution is $\frac{k^2 \cdot d_j}{k^2+d_j}$ times less than a standard convolution while having similar performance \cite{14}. Each "bottleneck" block consists of a $1 \times 1$ convolution to expand the number of channels by a factor of $t$, and a depthwise separable convolution of stride $s$ and output channels $c$. Multiple fully connected layers were used to achieve the two desired outputs of the network: the classification and the endpoints of the zebra crossing. Compared to MobileNet v2, LYTNet was adapted for a larger input of $768 \times 576 \times 3$ in order for the pedestrian traffic lights to retain a certain degree of clarity. We used a $2 \times 2$ max-pool layer after the first convolution to decrease the size of the output and thus increase the speed of the network. LYTNet also features significantly fewer bottleneck layers (10 vs 17) compared to MobileNet v2 \cite{14}. Table 2 shows the detailed structure of our network. 

During training, we used the Adam optimizer with momentum $0.9$ 
\begin{table}
\caption{Structure of Our Network}\label{tab2} 
\begin{tabular*}{\textwidth}{c @{\extracolsep{\fill}} ccccc}
\specialrule{.1em}{.05em}{.05em} 
Input & Operator & $t$ & $c$ & $n$ & $s$\\
\specialrule{.1em}{.05em}{.05em} 
$768 \times 576 \times 3$ & conv2d $3 \times 3$ & - & 32 & 1 & 2\\
\newline
$384 \times 288 \times 32$ & maxpool $2 \times 2$ & - & - & 1 & -\\
\newline
$192 \times 144 \times 32$ & Bottleneck & 1 & 16 & 1 & 1\\
\newline
$192 \times 144 \times 16$ & Bottleneck & 6 & 24 & 1 & 2\\
\newline
$96 \times 72 \times 24$ & Bottleneck & 6 & 24 & 2 & 1\\
\newline
$96 \times 72 \times 24$ & Bottleneck & 6 & 32 & 1 & 2\\
\newline
$48 \times 36 \times 32$ & Bottleneck & 6 & 64 & 1 & 2\\
\newline
$24 \times 18 \times 64$ & Bottleneck & 6 & 64 & 2 & 1\\
\newline
$24 \times 18 \times 64$ & Bottleneck & 6 & 96 & 1 & 1\\
\newline
$12 \times 9 \times 160$ & Bottleneck & 6 & 160 & 2 & 1\\
\newline
$12 \times 9 \times 160$ & Bottleneck & 6 & 320 & 1 & 1\\
\newline
$12 \times 9 \times 320$ & conv2d $1 \times 1$ & - & 1280 & 1 & 1\\
\newline
$12 \times 9 \times 1280$ & avgpool $12 \times 9$ & - & 1280 & 1 & -\\
\newline
1280 & FC & - & 160 & 1 & -\\
160 & FC & - & 5 & 1 & -\\
1280 & FC & - & 80 & 1 & -\\
\newline
80 & FC & - & 4 & 1 & -\\
\specialrule{.1em}{.05em}{.05em} 
\end{tabular*}
\end{table}
\begin{figure}
\includegraphics[width=\textwidth]{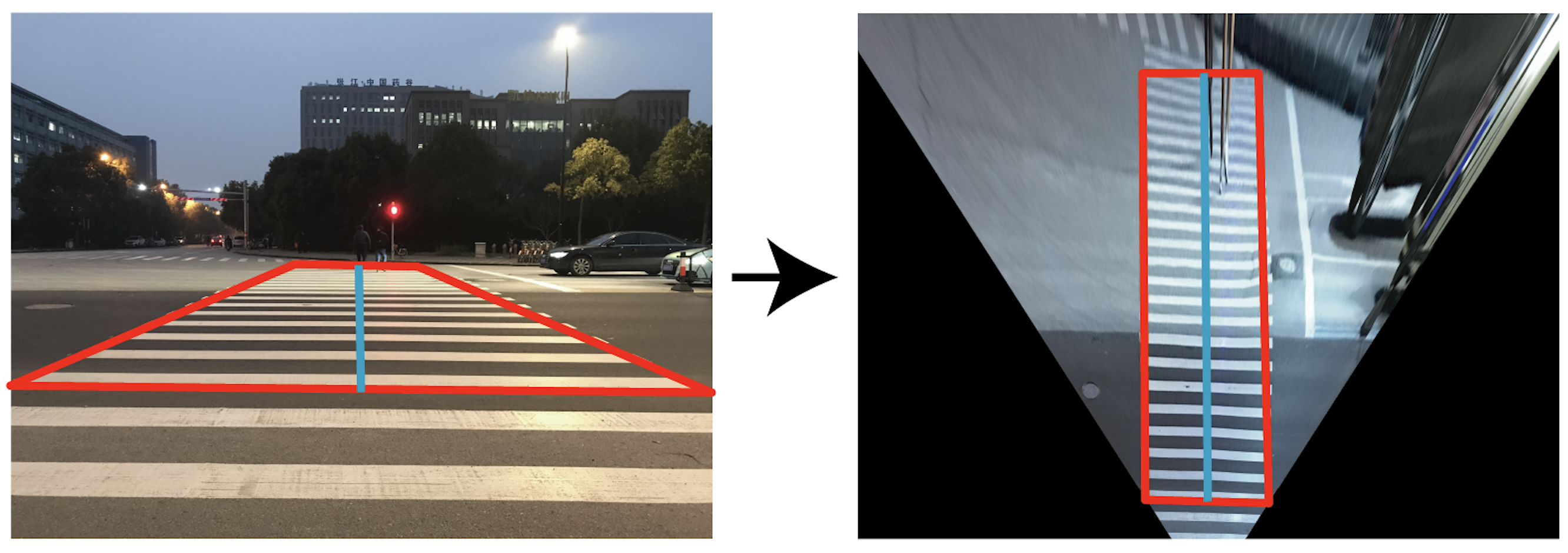}
\caption{The image on the left is the base image that was taken perpendicular to the zebra crossing and positioned in the center of the crossing, at a camera height of 1.4 m. Using our matrix, each point in the base image is mapped to a new point, creating the birds-eye image on the right. We can see that the zebra crossing is bounded by a rectangle with a midline centered and perpendicular to the x-axis.}\label{figure3}
\end{figure}
and initial learning rate of $0.001$. The learning rate was decreased by a factor of 10 at 150, 400, and 650 epochs, with the network converging at around 800 epochs. We used a combination of cross-entropy loss (for image classification to calculate the loss for classification) and mean-squared-error loss (for regression to calculate the loss for direction) function is defined as:
\begin{equation}
L = \omega \cdot MSE + (1-\omega) \cdot CE + \lambda \cdot R(W)
\end{equation}
in which $R(W)$ is L-2 regularization. We used the value $\omega = 0.5$ during training. 

\subsection{Conversion of 2D Image Coordinates to 3D World Coordinates}
The predicted endpoints output from the network are assumed to be accurate in regards to the 2D image. However, the appearance of objects and the zebra crossing in the image plane is an incorrect representation of the position of objects in the 3D world. Since the desired object, the zebra crossing, is on the ground, it has a fixed z-value of $z=1$, enabling the conversion of a 2D image to a 2D birds-eye perspective image to achieve the desired 3D real-world information of the zebra crossing.

On our base image in Figure 2, we define four points: (1671,1440), (2361,1440), (4032,2171), (0,2171) and four corresponding points in the real world: (1671,212), (2361,212), (2361,2812), (1671,2812), with the points defined on the xy-plane such that $0 \leq x < 4032$ and $0 \leq y < 3024$. The matrix
$$
\begin{bmatrix}
-1.17079727\cdot10^{-1} & -1.56391162\cdot10^{0} & 2.25203273\cdot10^{3}\\
0 & -2.59783431\cdot10^{0} & 3.71606050\cdot10^{3}\\
0 & -7.75749810\cdot10^{-4} & 1.00000000\cdot10^{0}
\end{bmatrix}
$$
maps each point on the image $(x,y,1)$ to its corresponding point in the real-world. Assuming a fixed height, and a fixed angle around the transverse and longitudinal axes, the matrix will perfectly map each point on the image to the correct birds-eye-view point. Though this is not the case due to varying heights and angles around the transverse axis, the matrix provides the rough position of zebra crossing in the real world, which is sufficient for the purposes of guiding the visually impaired to a correct orientation. 
\subsection{Mobile Application}
As a proof of concept, an application was created using Swift. LYTNet is deployed in the application. Additional post-processing steps are implemented in the application to increase safety and convert zebra crossing data into information for the visually impaired. Accordingly, the softmax probabilities of each class is stored in phone memory, and the probabilities are averaged over five consecutive frames. Since countdown\_blank and countdown\_green represent the same mode of traffic light - a green light that has numbers counting down - the probabilities of either class are added together. A probability threshold of 0.8 is set for the application to output a decision. This is used to prevent a decision from being made before or after the pedestrian traffic light changes color. If one frame of the five frame average is different, the probability threshold would not be reached. Users will be alerted by a choice of beeps or vibrations whenever the five-frame average changes to a different traffic light mode. The average of the endpoint coordinates is also taken over five consecutive frames to provide more stable instructions for the user. The direction is retrieved from the angle of the direction vector in the birds-eye perspective. 

A threshold of $10\degree$ was set for $\Delta\theta$ before instructions are output to the user. If $\Delta\theta < -10\degree$ then an instruction for the user to rotate left is output, and if $ \Delta\theta > 10\degree $ an instruction for the user to rotate right is output. The $x$-intercept of the line through the start and end-points is calculated with:
\begin{equation}
    x_{int}=\frac{x_1y_2-x_2y_1}{y_2-y_1}.
\end{equation}
For an image with width $w$ and midline at $(w-1)/2$, if $x_{int} > (w-1)/2 + w\cdot0.085$, instructions 
\begin{figure}
\includegraphics[width=\textwidth]{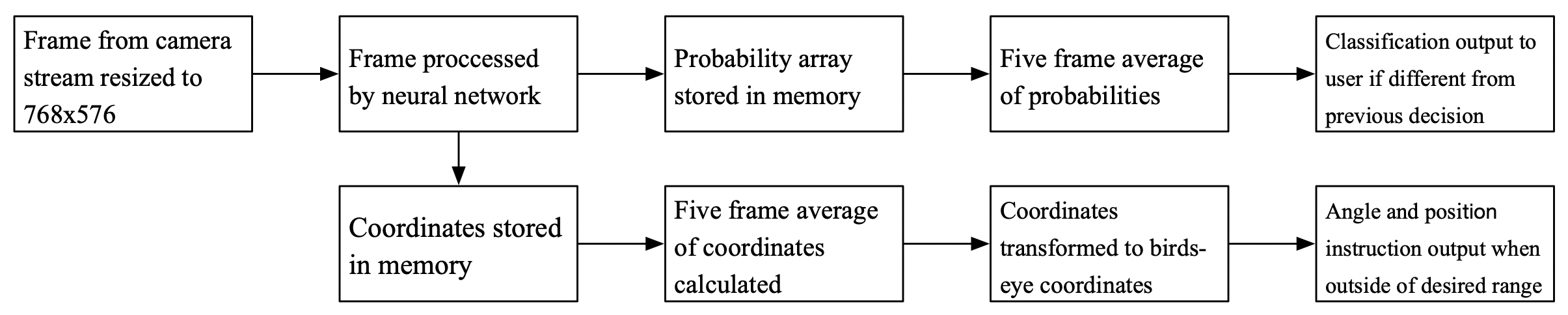}
\caption{Our application continuously iterates through this flow chart at 20fps.} \label{figure4}
\end{figure}
\begin{figure}
\includegraphics[width=\textwidth]{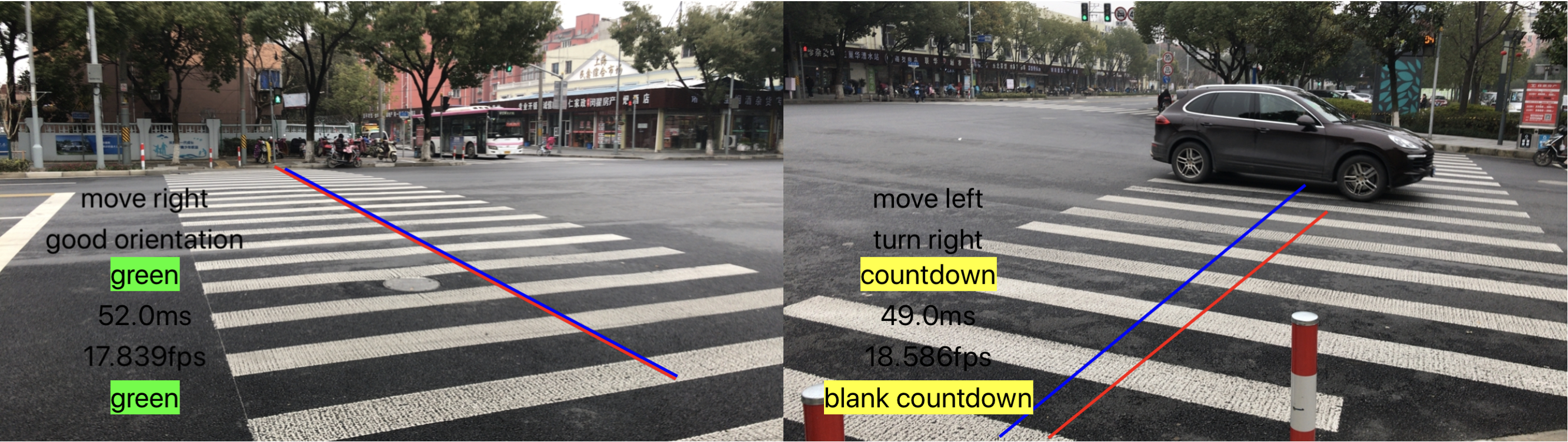}
\caption{Sample screenshots from our demo application. In order from top to bottom is the: position instruction, orientation instruction, 5-frame average class, delay, frame rate, and current detected class. The blue line is the direction vector for the specific frame , and the red line is the five-frame average direction vector.} \label{figure5}
\end{figure}
are given to move left, and if $x_{int} < (w-1)/2 - w\cdot0.085$, instructions are given to move right. In our defined area of the zebra crossing in transformed base image, the edges of the zebra crossing are within $8.5\%$ of the midline. With a constant width for the zebra crossing, if $x_{int}$ is outside of the $8.5\%$ range, the user will be outside of the zebra crossing. Refer to Figure 3 for a flow chart of the demo application and Figure 4 for a screenshot of our demo application. 
\section{Experiments}
We trained our network using 3456 images from our dataset and 864 images for validation \cite{2}. Our testing dataset consists of 739 images. 
The width multiplier changes the number of output channels at each layer. A smaller width multiplier decreases the number of channels and makes the network less computationally expensive, but sacrifices accuracy. As seen in Table 3, networks using a higher width multiplier also have a lower accuracy due to overfitting. We performed further testing using the network with width multiplier 1.0, as it achieves the highest accuracy while maintaining near real-time speed when tested on an iPhone 7. The precisions and recalls of countdown\_blank and none are the lowest out of all classes, which may be due to the limited number of training samples
\begin{table}
\caption{Comparison of Different Network Widths}\label{tab3} 
\begin{tabular*}{\textwidth}{c @{\extracolsep{\fill}} ccccc}
\specialrule{.1em}{.05em}{.05em} 
Width & Accuracy (\%) & Angle Error (degrees) & Start-point Error & Frame Rate (fps)\\
\specialrule{.1em}{.05em}{.05em} 
1.4 & 93.50 & 6.80 & 0.0805 & 15.69\\
\newline
1.25 & 92.96 & 6.73 & 0.0810 & 17.19\\
\newline
1.0 & 94.18 & 6.27 & 0.0763 & 20.32\\
\newline
0.9375 & 93.50 & 6.44 & 0.0768 & 21.69\\
\newline
0.875 & 93.23 & 7.08 & 0.0854 & 23.41\\
\newline
0.75 & 92.96 & 7.16 & 0.0825 & 24.33\\
\newline
0.5 & 89.99 & 7.19 & 0.0853 & 28.30\\
\specialrule{.1em}{.05em}{.05em} 
\end{tabular*}
\end{table}
\begin{table}
\caption{Precision and Recalls by Class}\label{tab4} 
\begin{tabular*}{\textwidth}{c @{\extracolsep{\fill}} ccccc}
\specialrule{.1em}{.05em}{.05em} 
 & Red      & Green     & Countdown Green & Countdown Blank & None    \\
\specialrule{.1em}{.05em}{.05em} 
Precision & 0.97 & 0.94 & 0.99 & 0.86 & 0.92\\
\newline
Recall & 0.96 & 0.94 & 0.96 & 0.92 & 0.87\\
\newline
F1 Score & 0.96 & 0.94 & 0.97 & 0.89 & 0.89\\
\specialrule{.1em}{.05em}{.05em} 
\end{tabular*}
\end{table}
for those two classes (Table 4). However, the precision and recall of red traffic lights, the most important class, is greater than 96\%.

When the zebra crossing is clear/unblocked, the angle error, startpoint, and endpoint errors are significantly better than when it is obstructed (Table 5). For an obstructed zebra crossing, insufficient information is provided in the image for the network to output precise endpoints. 

Figure 5 shows various outputs of our network. In (A), the network correctly predicts no traffic light despite two green car traffic lights taking a prominent place in the background, and is able to somewhat accurately predict the coordinates despite the zebra crossing appearing faint. In (B), the model correctly predicted the class despite the symbol being underexposed by the camera. (C) and (D) show examples of the model correctly predicting the traffic light despite rainy and snowy weather. (B), (C), and (D) all show the network predicting coordinates close to the ground truth. 

To prove the effectiveness of LYTNet, we retrained it using only red, green, and none class pictures from our own dataset and tested it on the PTLR dataset \cite{5}. Due to the small size of the PTLR training dataset, we were unable to perform further training or fine-tuning using the dataset without significant overfitting. Using the China portion of the PTLR dataset, we compared our algorithm with Cheng et al.'s algorithm, which is the most recent attempt for pedestrian traffic light detection to our knowledge. 

LYTNet was able to outperform their algorithm in regards to the F1 score, despite the disadvantage of insufficient training data from the PTLR dataset to train our network (Table 6). Furthermore, LYTNet provides additional information about the direction of the zebra crossing, giving the visually impaired a more comprehensive set of information for crossing the street, and outputs information regarding 4 different modes of traffic lights rather than only 2. We also achieve a similar frame rate to Cheng et al.'s algorithm, which achieved a frame rate of 21, albeit on a different mobile device. 
\begin{table}
\caption{Comparison of Network Performance on Clear and Obstructed Zebra Crossings}\label{tab5} 
\begin{tabular*}{\textwidth}{c @{\extracolsep{\fill}} ccccc}
\specialrule{.1em}{.05em}{.05em}
 & Number of Images & Angle Error & Startpoint Error & Endpoint Error  \\
\specialrule{.1em}{.05em}{.05em} 
Clear & 594 & 5.86 & 0.0725 & 0.476\\
\newline
Obstructed & 154 & 7.97 & 0.0918 & 0.0649\\
\newline
All & 739 & 6.27 & 0.0763 & 0.0510\\
\specialrule{.1em}{.05em}{.05em} 
\end{tabular*}
\end{table}
\begin{figure}
\includegraphics[width=\textwidth]{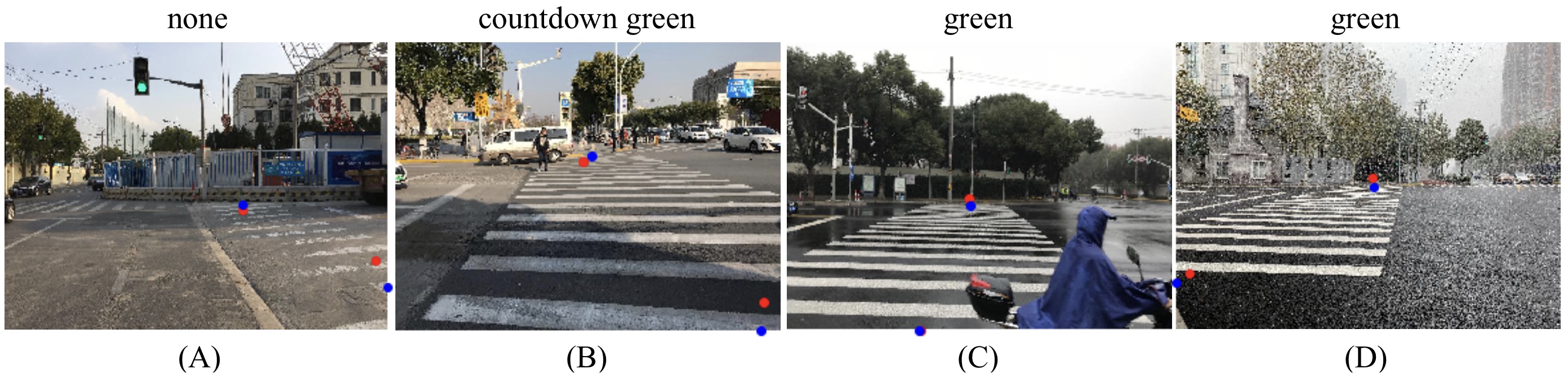}
\caption{Example correct outputs from our neural network. The class is labelled on top of each image. Blue dots are ground truth coordinates and red dots are predicted coordinates.} \label{new}
\end{figure}
\section{Conclusion}
In this paper, we proposed LYTNet, a convolutional neural network that uses image classification to detect the color of pedestrian traffic lights and to provide the direction and position of the zebra crossing to assist the visually impaired in crossing the street. LYTNet uses techniques taken from MobileNet v2,and was trained on our dataset, which is one of the largest pedestrian traffic light datasets in the world \cite{2}. Images were captured at hundreds of traffic intersections within Shanghai at a variety of different heights, angles, and positions relative to the zebra crossing.

Unlike previous methods that use multiple steps like detecting candidate areas, LYTNet uses image classification, a one-step approach. Since the network can learn features from an entire image rather than only detecting the pedestrian traffic light symbol, it has the advantage of being more robust in cases such as images with multiple pedestrian traffic lights. With sufficient training data, the network can draw clues from the context of an image along with the traffic light color to reach the correct prediction.

Additionally, LYTNet provides the advantage of being more comprehensive than previous methods as it classifies the traffic light between five total classes compared to 3 or 4 in previous attempts. Furthermore, our network is also capable of outputting zebra crossing information, which other methods do not
\begin{table}
\caption{Precision and Recall of Our Network and Cheng et al.'s Algorithm}\label{tab6} 
\begin{tabular*}{\textwidth}{c @{\extracolsep{\fill}} ccccc}
\specialrule{.1em}{.05em}{.05em} 
 & & Our Network & Cheng et al.'s Algorithm \\
\specialrule{.1em}{.05em}{.05em} 
Red & Recall & \textbf{92.23} & 86.43\\
 & Precision & 96.24 & \textbf{96.67}\\
 & F1 Score & \textbf{94.19} & 91.26\\
\newline
Green & Recall & \textbf{92.15} & 91.30\\
& Precision & \textbf{98.83} & 98.03\\
& F1 Score & \textbf{95.37} & 94.55\\
\specialrule{.1em}{.05em}{.05em} 
\end{tabular*}
\end{table}
provide. Thus, LYTNet elegantly combines the two most needed pieces of information without requiring two separate algorithms. Furthermore, our network is able to match the performance of the algorithm proposed by Cheng et al.

In the future, we will improve the robustness of our deep learning model through the expansion of our dataset, for further generalization. For the two classes with the least data, none and countdown\_blank, additional data can greatly improve the precisions and recalls. Data from other areas around the world can also be collected to separately train the network to perform optimally in another region with pedestrian traffic lights with differently shaped symbols. Our demonstration mobile application will be further developed into a working application that converts the output into auditory and sensory information for the visually impaired.
\section{Acknowledgements}
We would like to express our sincerest gratitude to Professor Chunhua Shen, Dr. Facheng Li, and Dr. Rongyi Lan for their insight and expertise when helping us in our research.

{\RaggedRight

}


\begin{thebibliography}{8}
\bibitem{1}
    Blaauw, F., Krieke, L., Emerencia, A., Aiello, M., Jonge, P.: Personalized advice for enhancing well-being using automated impulse response analysis – aira (06 2017)
\bibitem{2}
    Samuel, Y., Heon, L., John, K.: Ptl-dataset. \url{https://github.com/samuelyu2002/PTL-Dataset} (04 2019)
\bibitem{3}
    Mascetti, S., Ahmetovic, D., Gerino, A., Bernareggi, C., Busso, M., Rizzi, A.: Robust traffic lights detection on mobile devices for pedestrians with visual impairment. Computer Vision and Image Understanding \textbf{148} (12 2015). \doi{10.1016/j.cviu.2015.11.017}
\bibitem{4}
    Barlow, J., Bentzen, B., Tabor, L.: Accessible Pedestrian Signals. National Cooperative Highway Research Program (2003)
\bibitem{5}
    Cheng, R., Wang, K., Yang, K., Long, N., Bai, J., Liu, D.: Real-time pedestrian crossing lights detection algorithm for the visually impaired. Multimedia Tools and Applications \textbf{77} (12 2017). \doi{10.1007/s11042-017-5472-5}
\bibitem{6}
    Almeida, T., Macedo, H., Matos, L.: A Traffic Light Recognition Device, pp. 369-375. Springer International Publishing, Cham (2018). \doi{10.1007/978−3-319−77028−4\_49}
\bibitem{7}
    Omachi, M., Omachi, S.: Traffic light detection with color and edge information. In: 2009 2nd IEEE International Conference on Computer Science and Information Technology. pp. 284–287 (2009). \doi{10.1109/ICCSIT.2009.5234518}
\bibitem{8}
    Gong, J., Jiang, Y., Xiong, G., Guan, C., Tao, G., Chen, H.: The recognition and tracking of traffic lights based on color segmentation and camshift for intelligent vehicles. In: 2010 IEEE Intelligent Vehicles Symposium. pp. 431–435 (2010). \doi{10.1109/IVS.2010.5548083}
\bibitem{9}
    Varan, S., Singh, S., Kunte, R.S., Sudhaker, S., Philip, B.: A road traffic signal recognition system based on template matching employing tree classifier. In: International Conference on Computational Intelligence and Multimedia Applications (ICCIMA 2007). vol. 3, pp. 360-365 (2008). \doi{10.1109/ICCIMA.2007.190}
\bibitem{10}
    Behrendt, K., Novak, L., Botros, R.: A deep learning approach to traffic lights: Detection, tracking, and classification. In: 2017 IEEE International Conference on Robotics and Automation (ICRA). pp. 1370-1377 (2017). \doi{10.1109/ICRA.2017.7989163}
\bibitem{11}
    Shioyama, T., Wu, H., Nakamura, N., Kitawaki, S.: Measurement of the length of pedestrian crossings and detection of traffic lights from image data. Meas. Sci. Technol \textbf{13}, 1450-1457 (09 2002). \doi{10.1088/0957-0233/13/9/311}
\bibitem{12}
    Ash, R., Ofri, D., Brokman, J., Friedman, I., Moshe, Y.: Real-time pedestrian traffic light detection. In: 2018 IEEE International Conference on the Science of Electrical Engineering in Israel (ICSEE). pp. 1-5 (2018). \doi{10.1109/ICSEE.2018.8646287}
\bibitem{13}
    Angin, P., Bhargava, B., Helal, S.: A mobile-cloud collaborative traffic lights detector for blind navigation. In: 2010 Eleventh International Conference on Mobile Data Management. pp. 396-401 (2010). \doi{10.1109/MDM.2010.71}
\bibitem{14}
    Sandler, M., Howard, A., Zhu, M., Zhmoginov, A., Chen, L.C.: Mobilenetv2: Inverted residuals and linear bottlenecks. In: 2018 IEEE/CVF Conference on Computer Vision and Pattern Recognition. pp. 4510-4520 (06 2018). \doi{10.1109/CVPR.2018.00474}
\bibitem{15}
    de Charette, R., Nashashibi, F.: Traffic light recognition using image processing compared to learning processes. In: 2009 IEEE/RSJ International Conference on Intelligent Robots and Systems. pp. 333-338 (2009). \doi{10.1109/IROS.2009.5353941}
\bibitem{16}
    Ivanchenko, V., Coughlan, J., Shen, H.: Crosswatch: A camera phone system for orienting visually impaired pedestrians at traffic intersections. In: Lecture Notes in Computer Science. vol. 5105, pp. 1122-1128. Springer Berlin Heidelberg (2008). \doi{10.1007/978-3-540-70540-6\_168}
\bibitem{17}
    Poggi, M., Nanni, L., Mattoccia, S.: Crosswalk recognition through point-cloud processing and deep-learning suited to a wearable mobility aid for the visually impaired. In: New Trends in Image Analysis and Processing – ICIAP 2015 Workshops. vol. 9281, pp. 282–289. Springer International Publishing (09 2015). \doi{10.1007/978-3-319-23222-5\_35}
\bibitem{18}
    Lausser, L., Schwenker, F., Palm, G.: Detecting zebra crossings utilizing adaboost. In: ESANN. pp. 535-540 (2008)
\bibitem{19}
    David Banich, J.: Zebra Crosswalk Detection Assisted By Neural Networks. Master’s thesis, California Polytechnic State University (2016)
\end{thebibliography}
\end{document}